\documentclass[10pt,twocolumn,letterpaper]{article}

\makeatletter
\@namedef{ver@everyshi.sty}{}
\makeatother
\usepackage[pagenumbers]{cvpr} 

\usepackage{graphicx}
\usepackage{amsmath}
\usepackage{amssymb}
\usepackage{booktabs}
\usepackage{multirow}
\usepackage{tikz}
\usepackage{pgfplots}
\usepackage{indentfirst}
\usepackage[title]{appendix}

\usepackage[pagebackref,breaklinks,colorlinks]{hyperref}

\usepackage[capitalize]{cleveref}

\crefname{section}{Sec.}{Secs.}
\Crefname{section}{Section}{Sections}
\Crefname{table}{Table}{Tables}
\crefname{table}{Tab.}{Tabs.}

\def\cvprPaperID{****} 
\def\confName{CVPR}
\def\confYear{2022}

\def\bW{\mathbf{W}}
\def\bX{\mathbf{X}}
\def\bA{\mathbf{A}}

\begin{document}

\title{Binary Neural Networks as a general-propose compute paradigm for on-device computer vision}

\author{
Guhong Nie$^{1}$, Lirui Xiao$^{1}$, Menglong Zhu$^{1}$, Dongliang Chu$^{1}$, \\
Yue Shen$^{1}$, Peng Li$^{1}$, Kang Yang$^{1}$, Li Du$^{2}$, Bo Chen$^{1}$\footnote{corresponding author} \\
$^1$ DJI Innovations Inc, Shenzhen, China \\
$^2$ School of Electronic Science and Engineering, Nanjing University, China \\
\texttt{\{neal.nie, lawson.xiao, menglong.zhu, daniel.chu\}@dji.com} \\
\texttt{\{yue.shen, lipper.li, kang.yang\}@dji.com, ldu@nju.edu.cn} \\
\texttt{platinum.chen@dji.com}
}

\maketitle

\begin{abstract}
   For binary neural networks (BNNs) to become the mainstream on-device computer vision algorithm, they must achieve a superior speed-vs-accuracy tradeoff than 8-bit quantization and establish a similar degree of general applicability in vision tasks. To this end, we propose a BNN framework comprising 1) a minimalistic inference scheme for hardware-friendliness, 2) an over-parameterized training scheme for high accuracy, and 3) a simple procedure to adapt to different vision tasks. The resultant framework overtakes 8-bit quantization in the speed-vs-accuracy tradeoff for classification, detection, segmentation, super-resolution and matching: our BNNs not only retain the accuracy levels of their 8-bit baselines but also showcase 1.3-2.4$\times$ faster FPS on mobile CPUs. Similar conclusions can be drawn for prototypical systolic-array-based AI accelerators, where our BNNs promise 2.8-7$\times$ fewer execution cycles than 8-bit and 2.1-2.7$\times$ fewer cycles than alternative BNN designs. These results suggest that the time for large-scale BNN adoption could be upon us.
\end{abstract}

\begin{figure*}
\centering
\includegraphics[width=1.0\textwidth]{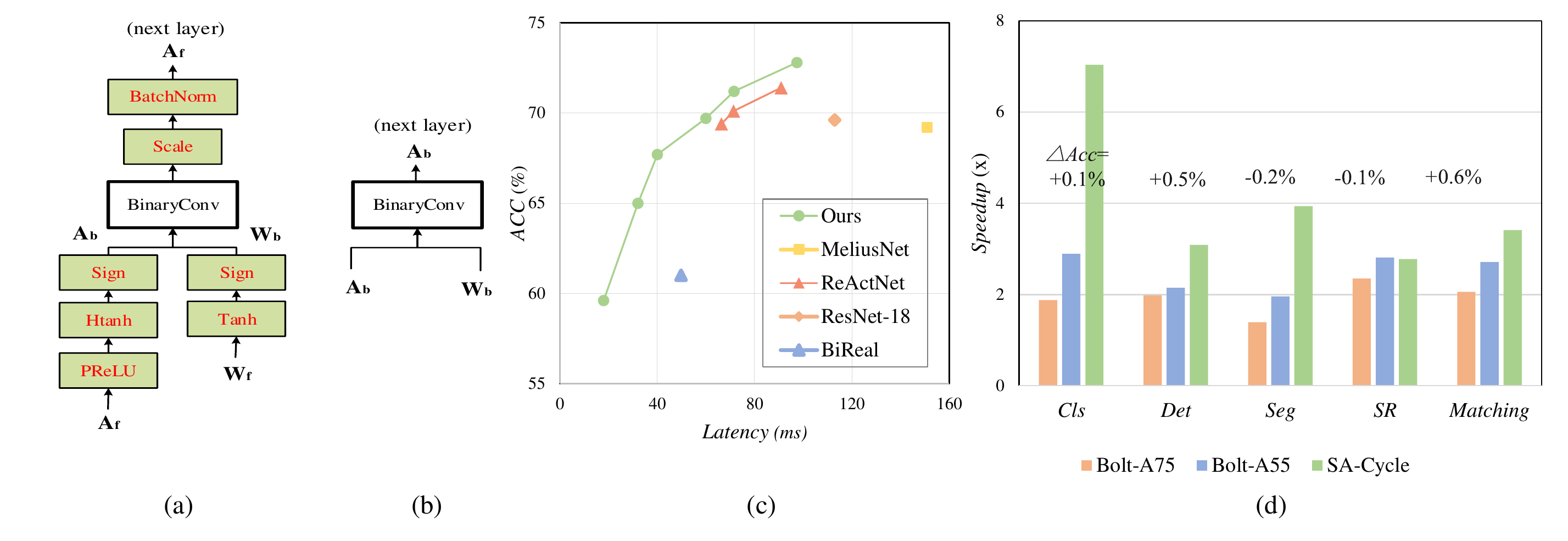}
\caption{{\it Speed-accuracy-tradeoff (SAT) and generality of BiNeal Networks}. (a) Quantization-aware training with over-parameterization. Auxiliary parameters and operators (green, shaded boxes) are introduced to enlarge network capacity, enabling BiNeal net to match 8-bit accuracy (Sec.~\ref{sec:over_parameter}). (b) Simplified inference form. Once trained, the auxiliary parameters and operators can be absorbed into the original parameters, yielding a inference form with just binary convolution (Sec.~\ref{sec:inference}). (c) BiNeal Net has a superior speed-vs-accuracy tradeoff in ImageNet classification than 8-bit and other binary networks. Latency measured using Bolt on Snapdragon 845 Cortex-A75@2.8GHz (Sec.~\ref{sec:exp_classification}). (d) Speedup of BiNeal networks over 8-bit on a variety of tasks at the similar accuracy or higher (Sec.~\ref{sec:generality}). Accuracy deltas are shown on top the bars. ``Bolt-X'' is latency measured on CPU X core, ``SA-Cycle'' are cycles required by a prototypical AI accelerator. }
\label{fig:main}
\end{figure*}

\section{Introduction}
\label{sec:intro}

Why aren't binary neural networks (BNNs) mainstream? BNNs promise increased compute intensity and reduced memory / data movement requirements, both of which are paramount for deep learning workloads in resource-constrained settings such as wearables, mobile phones and drones. BNNs have also witnessed a significant accuracy boost~\cite{lin2020rotated,xu2021recu} in ImageNet classification over the years since their inception in 2016\cite{courbariaux2016binarized}. Yet, despite the efficiency promise and the accuracy advancements, edge applications and deployment frameworks in industry \cite{gysel2018ristretto, jacob2018quantization, krishnamoorthi2018quantizing} predominantly prefer 8-bit or higher precision over BNNs. Some argues that this is in part still due to the lack of accuracy, as BNNs typically underperform floating-point models, with the exceptions of~\cite{liu2020reactnet,bethge2020meliusnet}. Nonetheless, we believe that the bottleneck to large-scale adoption lies instead in {\bf the speed-accuracy-tradeoff} (SAT) and {\bf generality}.

First, BNNs {\it do not need to} surpass 8-bit networks in accuracy alone; it may be sufficient to simply match 8-bit accuracy but {\it with a more economical resource budget} (e.g. latency). However, at the moment, accuracy is typically considered either in isolation or with metrics that are too coarse to reflect resource utilization on-device. A common competition is to ``close the gap'' between BNN and 8-bit networks {\it without altering model architecture}~\cite{lin2020rotated,xu2021recu,qin2020forward}. While this endeavor removes the burden of architecture design for 8-to-1-bit adoption, it ignores the resource disparity between the two and puts BNNs at a natural disadvantage. As a result, the current state-of-the-art BNNs \cite{lin2017towards, liu2018bi, zhuang2019structured, bethge2019binarydensenet} still lag behind 8-bit in accuracy.

It is shown that BNNs with modified network structure could obtain sizeable improvement in the SAT \cite{mishra2017wrpn, liu2018bi, liu2020reactnet, bethge2020meliusnet}. Yet, most prior work relies on the number of binary operations (BOPS) as a proxy for inference latency, which is overly simplistic for practical applications. BOPS only accounts for the computation latency from convolutions. It ignores significant latency contributors, including memory access \cite{ma2018shufflenet}, non-linearities, and cross-precision conversions. In addition, without a plausible hardware design, it is difficult and potentially misleading to compare BNN operations with 8-bit operations. For example, the commonly-used conversion of 1 8-bit OP $=64$ BOPS \cite{liu2020reactnet, bethge2020meliusnet} is found overly-optimistic in actual hardware instantiations \cite{yang2017bmxnet, zhang2019dabnn, geiger2020larq}.

Lastly, another roadblock for industry-wide adoption is the lack of confidence in {\it generality}. Recent quantization techniques \cite{bethge2019binarydensenet, liu2020reactnet, bethge2020meliusnet} are iterated on ImageNet classification, putting in question their transferability to other tasks. In particular, classification is ``lossy'' in nature:  it distils low-bandwidth categorical information from a high-bandwidth input image, discarding visual details in the process. Segmentation \cite{lim2017enhanced} and super-resolution\cite{noh2015learning}, on the other hand, typically has higher resolution in the output than the input, and may not tolerate information-loss the same way, as shown in \cite{ma2019efficient, xin2020binarized, jiang2021training}. It therefore remains to be seen how binary methods and network architectures favoured by  classification generalize to other tasks. 

Driven by these observations, we focus on proving BNN's superiority in SAT on actual hardware across vision applications. {\bf (1)} We binarize a ResNet block by enlarging its channels and injecting auxiliary parameters. This modification expands network capacity and helps the BNN to match 8-bit accuracy. Since the enlarged channels are binarized, they are less memory-intensive than 8-bit networks at the original channel count. {\bf (2)} During inference the BNN is transformed into a simple but mathematically equivalent inference form, where the auxiliary parameters are absorbed into the regular parameters. The bit-widths throughout the block are carefully balanced to both minimize cross-precision conversion and ensure sufficient representation capacity. The resultant block contains almost exclusively binary convolutions, with only one 4-bit addition in the end. No real-valued activation or nonlinearity is needed. The simplicity helps with reducing data movement and bandwidth, and leads to lower latency than other BNN designs and 8-bit networks using existing inference frameworks. It also enables convenient hardware acceleration, which we show using a cycle-calculation formula based on the standard systolic-array hardware design. {\bf (3)} As the proposed BNN block is a replacement for the commonly-used ResNet blocks, it can be transferred to most computer vision tasks without hyperparameter-tuning (although tuning can still be applied if desired). We show that a straight-forward transfer could deliver better SAT than current 8-bit networks.

Our main contributions are as follows:
\begin{itemize}
\item We propose a BNN structure, dubbed {\bf BiNeal networks} (with \underline{Bi}nary weights and \underline{N}o r\underline{eal}-valued activations), a BNN that obtains state-of-the-art accuracy among other BNNs, while enjoys a simple and parsimonious inference form for on-device deployment. 
\item We are {\bf the first to validate the generality of BNNs}. Without introducing additional hyperparameters, our BiNeal structure transfers to  classification, detection, segmentation, super-resolution and matching, and outperforms 8-bit networks and BNN alternatives in the SAT for each individual task.
\item We derive a metric to evaluate the inference speed of quantized networks in systolic-array based ASIC accelerators. The metric takes into account the latency incurred from by {\bf data movement}. It simultaneously enjoys more reliability than BOPS, and more convenience than manufacturing actual hardware.
\end{itemize}

\section{Related work}

\textbf{Binary Neural Networks}
BNNs are introduced by~\cite{courbariaux2016binarized} where weights and activations are only $+1$ or $-1$. This field is popularized by earlier attempts such as XNOR\cite{rastegari2016xnor}, ABC-Net\cite{lin2017towards}, and Bi-Real\cite{liu2018bi}. Recent work fall into two categories. One focuses on improving quantization algorithms based on the Bi-Real\cite{liu2018bi} structure, as exemplified by RBNN\cite{lin2020rotated}, ReCU\cite{xu2021recu}, and IR-Net\cite{qin2020forward}. RBNN\cite{lin2020rotated} introduces rotation matrices and angular biases to reduce the quantization loss, ReCU\cite{xu2021recu} introduces rectified clamp units to revive the "dead weights" for the purpose of reducing quantization error. Yet, these methods do not modify the downsampling operations in Bi-Real networks, which are conducted in floating-point and could incur large memory access.

The other cateogry of methods modify the network structure, as in GroupNet\cite{zhuang2019structured}, ReActNet\cite{liu2020reactnet}, BinaryDenseNet\cite{bethge2019binarydensenet}, and MeliusNet\cite{bethge2020meliusnet}. The modifications enable BNNs to outperform 8-bit networks in accuracy: ReActNet surpasses ResNet18 accuracy\cite{he2016deep}, while MeliusNet overtakes MobileNet\cite{howard2017mobilenets}. Nonethelss, it is unclear how the modified BNNs compare with 8-bit networks in on-device inference, as only model size and BOPS are considered for efficiency evaluation.

One notable exception is WRPN~\cite{mishra2017wrpn}. It not only matches full-precision accuracy by enlarging its network structure, but also provides hardware instantiations to showcase the efficiency win. We improve upon their work in three aspects: (i) higher accuracy with the BiNeal structure, (ii) a latency measurement formula that is more realistic than BOPS and does not require publishing ASIC design and synthesis for community adoption, (iii) generalization analysis on non-classification workloads.

\textbf{Application outside of classification} While the vast majority of modern BNN approaches report results in ImageNet classification, a few have also started experimenting on {\it object detection and tracking}\cite{sun2018fast,liu2019rbcn,yang2019binarized,wang2020bidet}. Yet, most attempts either report results on small-scaled datasets. The only exception is BiDet\cite{wang2020bidet}, which fails to bridge the gap with full-precision models on the standard COCO dataset.

{\it Single image super-resolution} has seen a recent surge in binarization efforts, but these efforts are still balancing between simplicity and accuracy.\cite{xin2020binarized} introduces a simple strategy to replace regular convolutions with binary convolutions, but at the expense of noticeable performance degradation from full-precision.\cite{ma2019efficient} only binarizes weights and \cite{huang2021binarizing} introduces scaling factors during inference, both of which sacrifices inference efficiency. BAM\cite{xin2020binarized} and IBTM\cite{jiang2021training} obtain decent SAT with task-specific network modifications, which is not available for other tasks.

BNN applications on {\it Segmentation} have been limited. GroupNet\cite{zhuang2019structured} combines BNNs with a segmentation-specific modification to match floating-point performance in PASCAL VOC segmentation, yet its basic version still is underperforming in classification by a large margin, thus lacking generality.

\textbf{BNN Inference frameworks}
There are multiple inference frameworks~\cite{zhao2018bitstream,hu2018bitflow,geiger2020larq,yang2017bmxnet,zhang2019dabnn,chen2020phonebit} that can export a trained BNN for on-device excution.
Among them, BitStream\cite{zhao2018bitstream} and BitFlow\cite{hu2018bitflow} are close-sourced; BMXNet\cite{yang2017bmxnet}'s 1-bit inference speed is slower than floating-point. 

There are several open-sourced BNN inference framework on ARM, such as daBNN \cite{zhang2019dabnn}, Larq\cite{geiger2020larq}, and Bolt\footnote{https://github.com/huawei-noah/bolt}.
DaBNN proposed an upgraded bit-packing scheme and several speed-up and memory
refinement strategies. Larq extends TensorFlow and TensorFlow Lite, and optimizes the implementations of binary operation at assembly level. PhoneBit~\cite{chen2020phonebit} is a GPU-accelerated BNN inference engine for Android-based mobile devices. Bolt optimizes binary convolution using tilegemm, and it is the fastest binary inference framework on ARM so far.
Limited by the scarsity in 1-bit hardware, these inference frameworks have to be shoehorned into existing 8-bit hardware, unable to utilize BNNs to their full potential. Vice versa, BNNs are not designed with these hardware in mind, which means that the architecture choices, such as channel counts, connectivity and operators, may not be aligned with hardware's preference. As a result, speedups are often less impressive for whole networks on actual devices than for an individual convolution according to the theoretical estimates.

\textbf{Latency metric for BNNs}
One way to fully exploit the compute and memory efficiency for BNNs is through specialized accelerators. A systolic array is an accelerator architecture design found in popular hardware, e.g. Google's TPU\footnote{https://cloud.google.com/tpu/docs/system-architecture-tpu-vm}, Tesla's FSD \cite{talpes2020compute}, and MIT's Eyeriss\cite{chen2016eyeriss}. These design so far are based on floating-point or 8-bit convolutions.

Various efforts have been made on binary accelerators. FINN\cite{umuroglu2017finn} proposes a framework for binary neural networks inference on FPGA. ReBNN\cite{guan2019recursive} focuses on reducing memory usage when training BNN. WRPN~\cite{mishra2017wrpn} synthesizes an ASIC for multiple precisions including binary. Reproducing latency measurements on specialized ASICs are often expensive, hence we seek to derive a cycle estimation formula based on common features in these designs, yielding a convenient and hardware-grounded latency metric.

\section{BiNeal Network}
\label{sec:Binarization Quantization Inference}
BiNeal net is a combination of a quantization-aware training technique based on over-parameterization (Sec.~\ref{sec:over_parameter}), a simplified inference scheme (Sec.~\ref{sec:inference}) and a block structure to replace ResNet (Sec.~\ref{sec:EfficientInferenceBlock}). We introduce each component below.

\begin{figure}
\centering
\includegraphics[width=\linewidth]{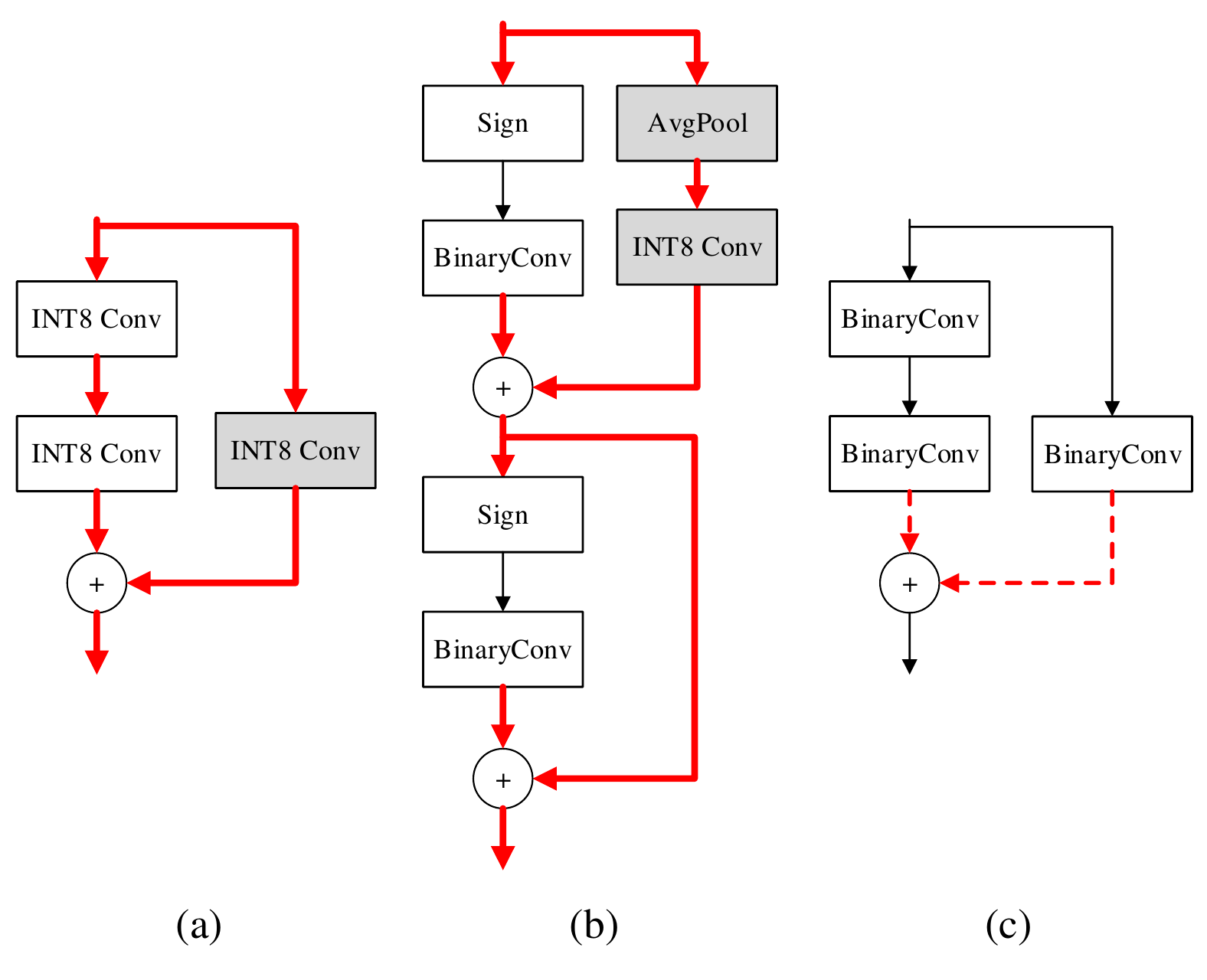}
\caption{Diagrams for (a) ResBlock, (b) Bi-Real block and (c) the proposed BiNeal block. Grey boxes indicate additional layers for down-sampling blocks. Solid red lines, dashed red lines and black lines represent 8-bit, 4-bit, and 1-bit dataflows, respectively.
}
\label{fig:binary_blocks}
\end{figure}

\subsection{Quantization-aware Training with Over-Parameterization}
\label{sec:over_parameter}
We start with the prototypical convolution operation. During training, we introduce additional parameters and structures that are apparently redundant but conducive to training performance. Over-parametrization appears in XNOR++, ReActNet and many others, but we are the first to absorb the auxiliary parameters in the bias term for efficient on-device deployment. Other BNN approaches only consider computational costs of the convolution operation, and ignore other operations that may introduce extra memory access and impact latency. In a similar fashion as\cite{ding2021repvgg}, the redundant parameterization can be absorbed into the conventional parameters during inference, at no cost to efficiency.

Let $\bW\in R^{N\times C\times K\times K}$, $\bX\in R^{B\times H\times W\times C}$ and $\bA=\text{Conv}(\bW, \bX)$ be the weight, input and output of the convolution, where $N, C, H, W, B, K$ are the output channel, input channel, height, width, batch size and kernel size (assumed square for simplicity).

\subsubsection{Weight Over-Parameterizaion}

Following DoReFa Net\cite{zhou2016dorefa}, we use floating-point weight $\bW_f$ for training, and approximately binarize them using $Tanh$ during the forward pass.
\begin{equation}
\begin{aligned}
{\bW_b} &= \text{Sign}(\text{Tanh}(\alpha \cdot \bW_f)) \label{eq:approx_W} \\
\bW_f &\approx \lambda \cdot {\bW_b} 
\end{aligned}
\end{equation}

where $\alpha$ and $\lambda$ are auxiliary parameters. $\alpha$ can assume different shapes ($\in R^N, \text{or}\ R^{N \times C},\text{or}\ R^{N \times C \times K \times K}$) to adjust the desired degree of over-parameterization, and the magnitude of each $\alpha$ coefficient controls the degree of sharpness of the approximation to binarization. $\lambda \in R^N$ is a per-output-channel scaling factor, introduced in XNOR\cite{rastegari2016xnor}, to compensate for the magnitude disparity between the unbounded weight $\bW_f$ and ${\bW_b}$ in ${\{-1, 1\}}$. 

Prior work fixes $\alpha$ at $1$\cite{zhou2016dorefa}, and set $\lambda$ to minimize the L1 difference between $\bW_f$ and ${\bW_b}$. Neither is proven ideal for end-to-end performance. Instead, we unfreeze $\alpha$ and $\lambda$ as free parameters, and train them along with the weights.

\subsubsection{Activation Over-Parameterization}
Since both the approximated weight and the convolution operation are in floating-point, the output is real-valued and must be binarized before being fed to the next layer. 
Following \cite{liu2018bi,liu2020reactnet}, we approximate the real-valued output $\bA_f$ using a binarized activation ${\bA_b}$ via a series of transformations:

\begin{equation}
\begin{aligned}
{\bA_b} &= \text{Sign}(\text{Htanh}(\text{PReLU}(\tau \bA_f + b_0) + b_1)) \label{eq:rpelu_kappa} \\
\bA_f &\approx \kappa \cdot {\bA_b} \\
\end{aligned}
\end{equation}
where $\tau, b_0, b_1 \in R^n$ and $\kappa \in R$ are auxiliary parameters, $\text{Htanh}$ is the hard-tanh function\cite{liu2018bi}. It clamps the input at $[-1, 1]$ during the forward pass, and uses sinosoids in the backward pass. PReLU\cite{liu2020reactnet} and Sign uses Straight-Through-Estimator to compute gradients.

At first sight the formulation above makes little sense, as 1) the transformations could be merged into an equivalent, per-channel thresholding operation, 2) the scale $\tau$ and biases $b_0$ are redundant given that $\bA_f$ is typically proceeded by a BatchNorm with its own learnable scale and bias. Nonetheless, as shown in~\cite{liu2020reactnet}, the over-parameterized formulation reshapes input distributions, which helps conditioning BNN training. Capitalizing on this phenomenon, we also assign scale and biases to the Sign function.

\subsection{Parameter-fused Inference}
\label{sec:inference}
The auxiliary parameters and non-linearities in training can be absorbed into a simple form during inference. Essentially, binarization only concerns with the {\it relative} value of two numbers, and disregards their magnitude. Exploiting this property, weight binarization in Eq.~\ref{eq:approx_W} is equivalent to:
\begin{equation}
{\bW_b} = \text{Sign}(\alpha) \cdot \text{Sign}(\bW_f) \label{eq:binary_W} 
\end{equation}

Similarly, for each output channel $n$, the activation binarization in Eq.~\ref{eq:rpelu_kappa} is equivalent to:
\begin{equation}
{\bA_b}(n) = \text{Sign}(\tau(n))\text{Sign}(\bA_f(n) - \theta(n)) \label{eq:binary_A}
\end{equation}
where $\theta(n) \in R$ is a threshold that depends on the $b_0(n), b_1(n)$ and $\tau(n)$ for channel $n$. This simplification owes to the fact that all transformations in Eq.~\ref{eq:rpelu_kappa} are monotonic given $b_0(n), b_1(n)$ and $\tau(n)$, thus one only needs to solve for the zero-point $\theta(n)$ to determine the sign of ${\bA_b}(n)$. The form and derivation can be found in (Appendix.~\ref{sec:apdix_theta}).

Finally, the convolutional output can be approximated by:
\begin{equation}
\text{Conv}(\bW_f, \bA_f) \approx (\kappa \cdot \lambda) \text{Conv}({\bW_b}, {\bA_b}) \label{eq:binary_Conv} \\
\end{equation}

\subsubsection{BatchNorm Fusion}

$(\kappa \cdot \lambda)\in R^N$ in Eq.~\ref{eq:binary_Conv} is a per-channel scaling factor that is carried into the next layer. Fortunately, they can be absorbed into the PReLU operation in the next layer, and thus do not need to be computed during inference. More generally, BatchNorms are also amenable to parameter-fusion. Let $\gamma, \beta \in R^N$ be the BatchNorm scale and bias:
\begin{equation}
\begin{aligned}
& PReLU(\tau \cdot BN(x) + b_0) + b_1 \\
& = PReLU(\tau \cdot (\gamma \cdot x + \beta) + b_0) + b_1 \\
& = PReLU((\tau \cdot \gamma) \cdot x + (\tau \cdot \beta + b_0)) + b_1 \label{eq:rpelu_kappa_batchnorm}
\end{aligned}
\end{equation}
Then the PReLU parameters are collapsed the same way as described in Eq.~\ref{eq:binary_A}.

To summarize, we take the over-parameterized form in Sec.~\ref{sec:over_parameter} and simplify it by fusing the auxiliary parameters.. The resultant inference form only involves bit-wise convolution and thresholding, as shown in Fig.~\ref{fig:main}.

\subsection{Efficient Inference Block Design}\label{sec:EfficientInferenceBlock}

After addressing binarization for a single convolution, we proceed to describe our strategy for binarizing a whole network.
We design a binary block based on BasicBlock\cite{he2016deep}, the basic building block for ResNet-18 and one of the most versatile blocks found in backbone designs across vision tasks. Similar to Bi-Real blocks\cite{liu2018bi}, our binary block serves as a drop-in replacement for BasicBlock with no additional hyperparameters.

Our design is illustrated in Fig.~\ref{fig:binary_blocks}(c). Specifically:
1) Regular convolutions in the BasicBlock are replaced with binary convolutions described in Sec~\ref{sec:inference}. The input and output channel counts are multiplied by a factor $m$ ($m=2$ in the experiments).
2) The skip connection is always equipped with a binary convolution, regardless of whether the layer downsamples or not.
3) Inputs into the elementwise-add are encoded in INT4. The sum is converted to 1-bit.

Compared to other BNN blocks such as Bi-Real, our block has the following properties. First, convolution inputs are all in 1-bit, eliminating the need for frequent precision conversion, and reducing the overall amount of data movement. Second, it does not contain PReLU, Sign or other nonlinearities. This eliminates the need for additional circuitry. Last, the skip connection is light-weight: both the additional convolution and the output are 1-bit, as opposed to being real-valued in Bi-Real. 

The simplicity of our block design delivers tangible latency benefits compared to alternative designs. Yet, it is not conveniently reflected in conventional complexity metrics such as BOPS or model size. In the section below, we describe a 1-bit ASIC accelerator design based on the classical systolic array, and derive a formula to calculate the compute cycles required for different blocks. 

\begin{figure*}[t]
  \centering
   \includegraphics[width=1.0\linewidth]{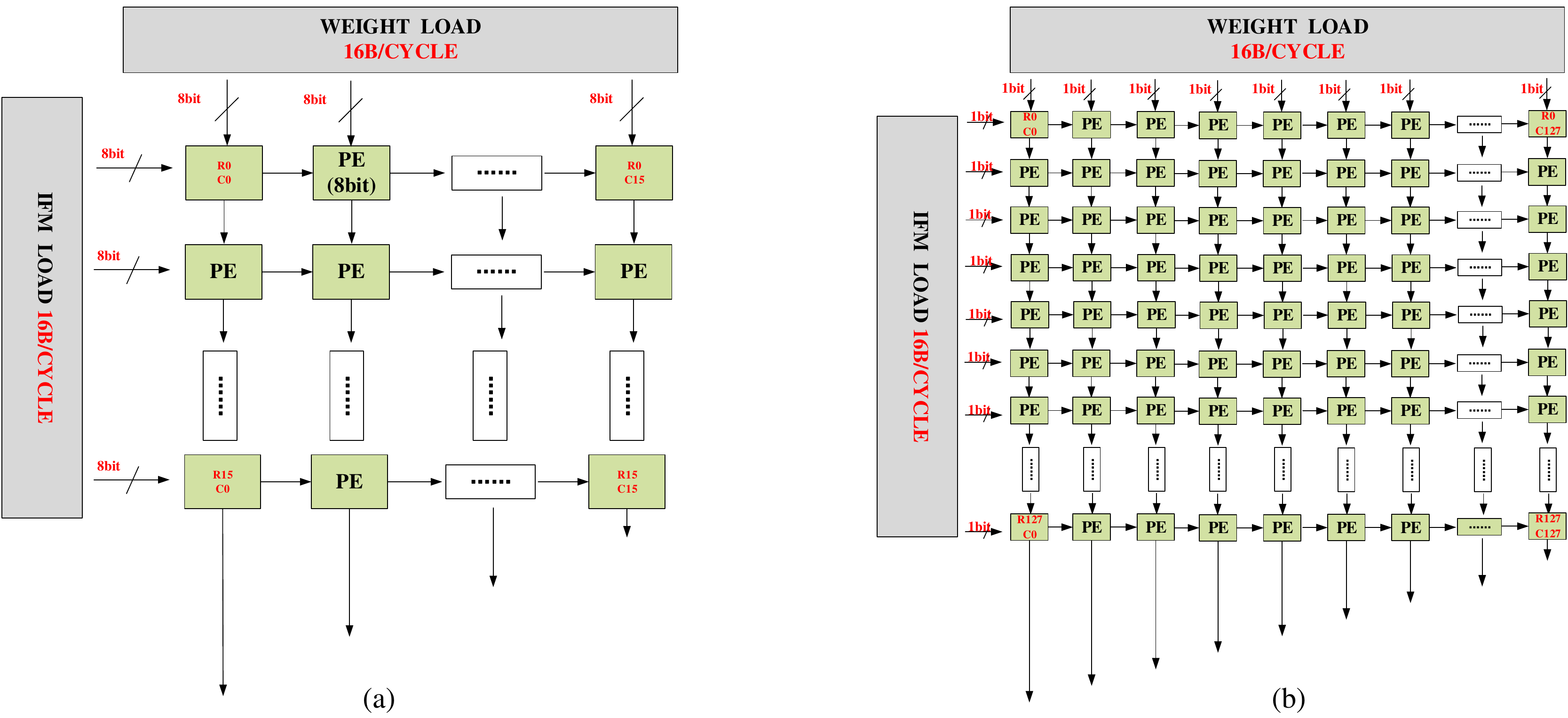}

   \caption{Example of Systolic Array Design.
   (a) A prototypical 8-bit array. (b) The modified 1-bit array with $8\times$ more columns and $8\times$ more rows. The input and weight loading bandwidth are held constant between the designs. }
   \label{fig:systolic_design}
\end{figure*}

\subsection{1-bit Systolic Array Design}\label{sec:SystolicArrayCycle}
We inherit a systolic array design in Fig.~\ref{fig:systolic_design}(a) for accelerating 8-bit convolution. It is prototypical to use a square array to ensure identical bandwidth for data-loads as well as writes without loss of generality. In turn, the input and output bandwidths (to and from SRAM) are typically identical. It is also common to store data in NHWC format (data arranged along each channel first).

To create a 1-bit systolic array, we replace the 8-bit PEs with 1-bit PEs. To utilize the full input bandwidth, we replace each 8-bit input with 8 1-bit input lines, essentially creating $8\times$ as many rows in the 1-bit design. As the array is square, the number of columns is also $8\times$. This translates to $64\times$ more PEs, and a theoretical $64\times$ speedup compared to 8-bit. PPA (performance or speed, power, and area) are the three key metrics for an ASIC. We use cycles as a proxy for speed of the ASIC, and derive a formula below for the total cycle count during inference of a given neural network. We also provides area and energy comparisons in (Appendix.~\ref{sec:apdix_area_energy}).

\subsection{Cycle Metric}
\label{sec:cycle}
In practice, speedup is often subject to factors such as array utilization, data movement, as well as the non-convolution parts of the workload. Based on the systolic-array design, we derive a set of formulae that describes the the number of cycles required to process a network. 

We assume that the weights are {\it not dynamically generated} during inference. This means that they are known in advance and can be pre-arranged to the desired NHWC format without incurring latency.

A benefit of the systolic-array design is that in-place operations can be pipelined (no additional cycles required) as long as they do not alter bit-depth. For example, BatchNorm and scaling can be considered cycle-free. As a result, to compute the cycles for blocks in Fig.~\ref{fig:binary_blocks}, one needs to consider convolution, bit-depth conversion, and elementwise-add. We briefly present their formulae below, and leave the derivations to the Appendix.~\ref{sec:cycle_formula}.

Let the array size be $S\times S$. For example, S=128 in Fig.~\ref{fig:binary_blocks}(a). Denote $M$ the input bandwidth (in bits per cycle), $P$ the size of Psum memory (i.e. the number of partial sums that can be stored on-chip), and $N,H,W,C$ represent the output channel, {\it output} height (note the difference from the previous notation), {\it output width}, and input channel, respectively.

\subsubsection{Convolution}
 The cycle count for the computational portion in a convolution is given by:
\begin{equation}
\text{T(*)} = \lceil \frac{N}{S} \rceil  \lceil \frac{W H}{P} \rceil S \label{instruction_cost}
\end{equation}

When the input and output have identical bit-depth (e.g. 1-bit in and 1-bit out), the cycles for the entire convolution is given by:
\begin{equation}
\text{T(Conv$_{1\rightarrow 1}$)} = W H K \lceil \frac{CK}{S} \rceil  \lceil \frac{N}{S} \rceil + \text{T(*)} \label{conv_cycle}
\end{equation}

When the output bit-depth $b$ is higher (e.g. 1-bit in and 8-bit out), the outgoing data will get congested. As input and output bandwidths are identical, higher output bit-depth translates to small number of results written per cycle. As a consequence, additional cycles, of number proportional to the output bit-depth $b$, are expensed in data movement:

\begin{equation}
\begin{aligned}
 \text{T(Conv$_{1\rightarrow b}$)} = W H K ( \lceil \frac{CK}{S} \rceil - 1 + b) 
  \lceil \frac{N}{S} \rceil + \text{T(*)} 
\end{aligned}
\label{bnn_conv_1imo}
\end{equation}

\subsubsection{Quantization and Sign}
One may need to re-quantize an input if its bit-depth does not match that of the PE (e.g. converting an real-valued input to 1-bit in Bi-Reall). The cycle cost is essentially identical to the cost of reading the input feature map:
\begin{equation}
 \text{T(read)} = B W H \lceil \frac{C}{S} \rceil S / M 
\label{quant_cycle}
\end{equation}

\subsubsection{Elementwise Operations}
 The write portion of elementwise-sum can be pipelined and thus omitted in cycle calculations. The cycle count is given by those of reading the two inputs:
\begin{equation}
 \text{T(+)} = 2 \text{T(read)}
\label{elt_cycle}
\end{equation}

So far, we have outlined all cycle calculations necessary for analysing blocks listed in Fig.~\ref{fig:binary_blocks}. Please refer to the (Appendix.~\ref{sec:cycle_formula} for an extensive list and derivations.

\subsubsection{Cycles for Common Network Blocks}
\label{sec:blocks}
We instantiate a typical systolic-array design using $S=128$, $P=1024$ and $M=128$. We use this setup to compute the cycles for network blocks in Fig.~\ref{fig:binary_blocks} as well as full networks in the experiments. 

The cycle counts for various block designs are shown in Tab.~\ref{tab:blocks_cycle}. BiNeal blocks are more amenable to acceleration despite having less BOPS with the other designs.

\begin{table}
  \centering
  \begin{tabular}{cc}
    \toprule
    Block & Cycle (K) \\
    \midrule
    ResBlock w/o downsample & 916.0 \\
    Bi-Real Block w/o downsample & 63.0 \\
    Ours w/o downsample & 32.5 \\
    ResBlock with downsample & 715.4 \\
    Bi-Real Block with downsample & 103.9 \\
    Ours with downsample & 26.3 \\
    \bottomrule
  \end{tabular}
  \caption{Cycles comparison with different block}
  \label{tab:blocks_cycle}
\end{table}

\section{Experiments}
Despite progress in quantized training for BNNs, there remains a discernible performance gap between BNNs and their floating-point (FP) counterparts. Similar to WRPN\cite{mishra2017wrpn}, BiNeal nets uniformly expand their constituent blocks by a channel multiplier. We compare BiNeal nets under various multiplier settings against FP, 8-bit and BNN models on ImageNet classification in Sec. \ref{sec:exp_classification}. We examine the latency on mobile devices and cycle metrics, showing superior SAT with BiNeal net's simple design. We then demonstrate BiNeal net's general applicability on object detection, semantic segmentation, single-frame super-resolution and image matching in Sec.\ref{sec:generality}.

\subsection{SAT on Classification}\label{sec:exp_classification}
We conduct classification experiments on ImageNet. Tab.\ref{tab:cycle_accuracy_display} displays the performance comparison among the proposed BiNeal nets, with  ResNet-18\cite{he2016deep},\cite{liu2018bi} (floating-point (FP) and 8-bit) and popular BNNs, namely Bi-Real\cite{liu2018bi}, ReActNet\cite{liu2020reactnet}, MeliusNet\cite{bethge2020meliusnet}, RBNN\cite{lin2020rotated} and ReCU\cite{xu2021recu}. 

All models are tested on the Snapdragon 845 CPU (Cortex-A75@2.8GHz). For each method we choose the best inference engine. Overall, FP and 8-bit models prefer TFLite whereas 1-bit models are consistently faster on Bolt. We systematically vary the multiplier to sweep a SAT curve. 

With a channel multiplier of $m=1.5$, our BiNeal net matches 8-bit ResNet-18 while  achieving a {\bf 1.9$\times$} speedup on Bolt and a {\bf 7.0$\times$} speedup measured by cycle metric. Our method is also {\bf 1.1$\times$} / {\bf 2.7$\times$} faster in latency / cycles than ReActNet-A, the most competitive BNN alternatives.

\begin{table}[ht]
  \centering
  \begin{tabular}{cccc}
    \toprule
    multiplier & Top-1 (\%) & latency (ms) & cycles (M)\\
    \midrule
    ResNet-18 FP       & 69.6    & 237.3 & - \\
    ResNet-18 (8bit)   & 69.6    & 112.93 & 7.47 \\
    ReActNet-A\cite{liu2020reactnet}  & 69.4    & 66.42 & 2.90 \\
    ReActNet-B\cite{liu2020reactnet}  & 70.1    & 71.37 & - \\
    ReActNet-C\cite{liu2020reactnet}  & 71.4    & 91.00 & - \\
    MeliusNet-42\cite{bethge2020meliusnet} & 69.2   & 151.01 & - \\
    MeliusNet-59\cite{bethge2020meliusnet} & 71.0   & 273.85 & - \\
    Bi-Real\cite{liu2018bi}		& 56.4    & 49.87 & 1.65 \\
    RBNN \cite{lin2020rotated} 		& 59.9 	& 49.87 & 1.65 \\
    ReCU \cite{xu2021recu} 			& 61.0 	& 49.87 & 1.65 \\
    ours-0.5x		& 50.9	& 13.19 & 0.74 \\
    ours-0.75x 		& 59.6	& 18.03 & 0.80 \\
    ours-1.0x 		& 65.0	& 32.05 & 0.84 \\
    ours-1.25x 		& 67.7 	& 40.09 & 1.00 \\
    \textbf{ours-1.5x} 		& \textbf{69.7} 	& \textbf{60.08} & \textbf{1.06} \\
    ours-1.75x 		& 71.2 	& 71.63 & 1.17 \\
    ours-2.0x 		& 72.8 	& 97.47 & 1.25 \\
    
    \bottomrule
  \end{tabular}
  \caption{Performance comparison of classification on ImageNet. Bolt latency and cycle estimation are based on the input with $224 \times 224 \times 3$.}
  \label{tab:cycle_accuracy_display}
\end{table}

\begin{table}[htbp]
  \centering
  \begin{tabular}{cccc}
    \toprule
    model & mAP & Bolt Latency (ms) & Cycle (G) \\
    \midrule
    CenterNet\cite{zhou2019objects}		& 28.1 	& -  & -\\
    CenterNet\cite{chen2019mmdetection} 		& 29.5	& 910.0 & - \\
    \textbf{CenterNet-ours} 		& \textbf{29.4}	& \textbf{722.57} & - \\
    \textbf{CenterNet-ours*} 		& \textbf{29.2} 	& \textbf{672.54} & - \\
    Faster-rcnn-fpn 		& 31.3 	& 1574.0 & 69.55 \\
    \textbf{ours} 		& \textbf{31.1} 	& \textbf{793.5} & \textbf{22.54} \\
    \bottomrule
  \end{tabular}
  \caption{Performance comparison of object detection on COCO. Bolt latency and cycle estimation are based on the input with $448 \times 672 \times 3$. CenterNet-ours* means that part of the head is binarized.}
  \label{tab:experiment_coco_cmp}
\end{table}

\subsection{Generality}\label{sec:generality}
We proceed to demonstrate BiNeal net's performance on common vision tasks. We deliberately avoid architecture specialization and choose one multiplier $m=2$ for all tasks. We use Cosine Annealing LR and WD$=$1E-5 for all tasks.While the performance per task may be further improved by structural and hyper-parameter tuning, the one-size-fits-all protocol seeks to establish a transferability baseline. This baseline not only helps with our generality assessment, but also mimics the practical setting of deploying models on novel tasks under time pressure.  

As 8-bit typically achieves less or equal accuracy than FP networks, we chivalrously assume that 8-bit could achieve FP accuracy in our comparison. All below baseline models are considered as 8-bit and the latency is tested on Bolt with Snapdragon 845 Cortex-A75@2.8GHz. We describe the setup and results for each task below.

\subsubsection{Detection}
Our detection experiment is conducted on COCO using the {\it mmdetection}\cite{chen2019mmdetection} implementation. We choose two popular model architectures: the anchor-free  centernet\cite{zhou2019objects} and the anchor-based faster-rcnn \cite{ren2015faster} with fpn\cite{lin2017feature}. The backbone for both model are ResNet-18, which we replace with BNNs. This allows us to reuse the binary classification models mentioned in Sec. \ref{sec:exp_classification} as pre-trained backbone initializations. We keep all the settings the same with that from origin FP network.

The results are shown in Tab.~\ref{tab:experiment_coco_cmp}.
For Centernet we report the mmdetection result\cite{chen2019mmdetection} (which is higher than the original paper\cite{zhou2019objects}). CenterNet-ours* refers to the model where both the backbone and head are binarized with the multiplier $m=2$. 

CenterNet with BiNeal backbones achieves comparable performance as the 8-bit model. Binarized CenterNet outperforms 8-bit model with {\bf 1.3$\times$} latency speedup. Binarized RCNN outperforms 8-bit model with {\bf 2.0$\times$} faster inference latency and {\bf 3.1$\times$} less cycles.

\subsubsection{Segmentation}
Our segmentation experiment is based on the CityScape\cite{chen2019mmdetection} dataset, implemented with mmsegmentation\cite{chen2019mmdetection}. Similar to detection, we binarize ResNet-18 backbones and pre-train them on ImageNet. We binarize FCN \cite{noh2015learning} using the proposed binary scheme and keep all the settings the same with that from original FP network. 

Tab. \ref{tab:experiment_cityscape_cmp} shows that our binarized model achieves comparable performance with the 8-bit model. Our binary model achieves {\bf 1.4$\times$} speedup on Bolt and {\bf 3.9$\times$} speedup on cycle metric.

\begin{table}[htbp]
  \centering
  \begin{tabular}{cccc}
    \toprule
    model & mIoU & Bolt Latency (ms) & Cycle (G) \\
    \midrule
    FCN	\cite{chen2019mmdetection} 	& \textbf{70.24} 	& 1230  & 81.16 \\
    \textbf{Ours} 		& 70.07	& \textbf{883.46} & \textbf{20.62} \\
    \bottomrule
  \end{tabular}
  \caption{Performance comparison of segmentation on cityscape. Bolt latency and cycle estimation are based on the input with $512 \times 1024 \times 3$.}
  \label{tab:experiment_cityscape_cmp}
\end{table}

\begin{table*}[ht]
\centering
\begin{tabular}{c|c|c|c|c|c|c}

\hline
\multirow{2}{*}{Method} & \multirow{2}{*}{scale} & \multirow{2}{*}{Param} & Set5      & Set14     & B100      & Urban100  \\
                        &                        &                        & PSNR/SSIM & PSNR/SSIM & PSNR/SSIM & PSNR/SSIM \\
\hline
EDSR                    & x2                     & 40.73M                 & 38.19/0.960           & 33.95/0.918    & 32.35/0.902  & 32.97/0.936      \\
EDSR\_16\_64            & x2                     & 1.370M                 & 37.81/0.959           & 33.34/0.913    & 32.04/0.898  & 31.46/0.922      \\
EDSR\_IBTM              & x2                     & 31.73M                 & 37.80/0.960           & \textbf{33.38/0.916}    & 32.04/0.898  & \textbf{31.49/0.922}      \\
EDSR\_ours               & x2                     & 5.039M                 & \textbf{37.85/0.959}           & 33.36/0.914    & \textbf{32.06/0.898}  & 31.46/0.923      \\
\hline
EDSR                    & x3                     & 40.73M                 & 34.68/0.928         & 30.53/0.844    & 29.26/0.809  & 28.81/0.868      \\
EDSR\_16\_64            & x3                     & 1.370M                 & 34.24/0.924         & 30.23/0.836    & 29.03/0.802  & 27.83/0.844      \\
EDSR\_IBTM              & x3                     & 31.73M                 & 34.10/0.924         & 30.11/0.838    & 28.93/0.801  & 27.49/0.839      \\
EDSR\_ours               & x3                     & 5.039M                 & \textbf{34.19/0.925}         & \textbf{30.23/0.838}    & \textbf{29.02/0.803}  & \textbf{27.80/0.848}      \\
\hline
EDSR                    & x4                     & 40.73M                 & 32.48/0.894         & 28.82/0.781    & 27.72/0.736  & 26.65/0.805      \\
EDSR\_16\_64            & x4                     & 1.370M                 & 32.14/0.888         & 28.56/0.773    & 27.56/0.728  & 25.89/0.778      \\
EDSR\_IBTM              & x4                     & 31.73M                 & 31.84/0.890         & 28.33/0.777    & 27.42/0.732  & 25.54/0.769      \\
EDSR\_ours               & x4                     & 5.039M                 & \textbf{31.94/0.887}         & \textbf{28.47/0.771}    & \textbf{27.49/0.726}  & \textbf{25.80/0.776}      \\
\hline
\end{tabular}
\caption{EDSR Result comparison. Models are evaluated on Set5\cite{bevilacqua2012low}, Set14\cite{zeyde2010single}, B100\cite{martin2001database}, Urban\cite{huang2015single}}
\label{tab:edsr_compare}
\end{table*}

\subsubsection{Super-resolution}
For single-frame super-resolution, we embed BNNs in EDSR\cite{lim2017enhanced}, which is one of the most iconic networks in the SR. The original network is extremely large, making channel expansion difficult. Instead, we binarize the smaller version of EDSR in the official release, which also matches the performance of the original. The smaller network differs in the following sense: (1) it uses 16 residual blocks instead of 32, (2) each block has 64 channels instead of 256, (3) we follow IBTM\cite{jiang2021training} and set the residual scale parameter to $1$ instead of $0.1$.
\begin{table}[htbp]
  \centering
  \begin{tabular}{ccc}
    \toprule
    model  & Bolt Latency (ms) & Cycle (G) \\
    \midrule
    EDSR\_16\_64 	& 9258.1   & 307.7 \\
    \textbf{Ours} 		& \textbf{3939.4} & \textbf{110.9} \\
    \bottomrule
  \end{tabular}
  \caption{Performance comparison of SR. Bolt latency and cycle estimation are based on the input with $192 \times 192 \times 3$.}
  \label{tab:experiment_sr}
\end{table}

We leave the first and last layers of the network in 8-bit, and replace all the residual blocks by BiNeal blocks with $m=2$. We follow the training settings of EDSR, except that we also follow\cite{jiang2021training} to normalize the input to $1$ instead of $255$.

We compare our method with previous SR networks, including EDSR\cite{lim2017enhanced} and EDSR-IBTM\cite{jiang2021training}. Tab. \ref{tab:edsr_compare} shows that our proposed binary EDSR networks have equivalent PSNR and SSIM score compared with the full precision model, but achieve {\bf 2.4$\times$} lower latency and {\bf 2.8$\times$} less cycles.

\subsubsection{Matching}
For image matching, we applied our binary scheme to R2D2\cite{revaud2019r2d2}. We binarize most the network and leave the last two output layers intact. L2-Net consists mostly of convolutions. Every two consecutive convolutions could be regarded as a BasicBlock without skip connections. Therefore, we use four BiNeal blocks with $m=2$ to replace eight floating-point convolution layers in L2-Net. We train the binary R2D2 model on Web image (W), Aachen day-time images (A) and Aachen optical flow pairs (F) and evaluate it on HPatches dataset. Mean Matching Accuracy (MMA) at an error threshold of 3px is the most popular accuracy metric for matching. Experimental results are shown in Tab. \ref{tab:experiment_matching}. 

Our binary R2D2 model achieves better matching performance than the 8-bit model, at the same time achieves {\bf 2.1$\times$}  speedup while inference using Bolt, and achieves {\bf 3.4$\times$} speedup evaluated by the cycle metric.

\begin{table}[htbp]
  \centering
  \begin{tabular}{cccc}
    \toprule
    model & MMA@3 & Bolt Latency (ms) & Cycle (G) \\
    \midrule
    R2D2 	& 0.686 	& 14942  & 906.7\\
    WRPN 	& 0.645	& 5973 & 180.6 \\
    \textbf{ours} 	& \textbf{0.692}	& \textbf{7262} & \textbf{265.7} \\
    \bottomrule
  \end{tabular}
  \caption{Performance comparison of MMA@3 on HPatches.  Bolt latency and cycle estimation are based on the input with $598 \times 796 \times 3$.}
  \label{tab:experiment_matching}
\end{table}

\section{Conclusion}\label{sec:conclusion}
We identified SAT superiority and generality as the two missing links to BNN's industrial success. Designed with inference efficiency in mind, our BNN managed to match 8-bit performance with greater frugality. This frugality was observed both in terms of direct latency measurements on mobile CPUs and theorectical cycle counts for prototypical AI accelerators. Moreover, the SAT advantage was observed across everyday vision workloads including classification, detection, segmentation, super-resolution and matching, delivering a consistent 1.3-2.4$\times$ speedup on ARM CPU. Cycle analysis reveals that the speedup could be 2.8-7.0$\times$ with a dedicated ASIC design. Importantly, our BNN framework adapts to these tasks without much need for hyperparameter-tuning. These results suggest that BNNs are general learners of computer vision tasks, and the opportune choice for replacing 8-bit as the default inference paradigm in resource-hungry scenarios.

{\small
\bibliographystyle{ieee_fullname}
\bibliography{egbib}
}

\clearpage
\begin{appendices}

\section{Fusion of Auxiliary Parameters} \label{sec:apdix_theta}

As shown in Eq.~\ref{eq:rpelu_kappa_batchnorm}, batchnorm can be absorbed into the PReLU operation in the next layer, so it is equivalent to Eq.~\ref{eq:rpelu_kappa}. Now, we derive Eq.~\ref{eq:binary_A} below.

We define $x$ as:
\begin{equation}
x = \tau \bA_f + b_0 \label{eq:tau_af}
\end{equation}

So Eq.\ref{eq:rpelu_kappa_batchnorm} can be defined as:

\begin{equation}
\begin{aligned}
& \bA_b = Sign(Htanh(PReLU(x) + b_1)) \\
& = Sign(PReLU(x) + b_1)
\label{eq:prelu_sign}
\end{aligned}
\end{equation}

Let $\alpha$ as the parameter of $PReLU$, $\alpha > 0, \alpha \in R^N$, so $PReLU$ can be defined as:
\begin{equation}
PReLU(x) = 
\begin{cases}
\alpha x & x \leq 0 \\
x & x > 0 \\
\end{cases}
\label{eq:prelu}
\end{equation}

We split Eq.\ref{eq:prelu_sign} in two parts.

Case $1$: $ b_1 \leq 0 $

\begin{equation}
\begin{aligned}
& \bA_b = Sign(PReLU(x) + b_1) \\
& = Sign(\alpha x + b_1) \\ 
& = \begin{cases}
-1 & x \leq -b_1 / \alpha \\
1 & x > -b_1 / \alpha \\
\end{cases}
\label{eq:prelu_alpha_case1}
\end{aligned}
\end{equation}

Case $2$: $ b_1 > 0$

\begin{equation}
\begin{aligned}
& \bA_b = Sign(PReLU(x) + b_1) \\
& = Sign(x + b_1) \\ 
& = \begin{cases}
-1 & x \leq -b_1 \\
1 & x > -b_1 \\
\end{cases}
\label{eq:prelu_alpha_case2}
\end{aligned}
\end{equation}

Considering Eq.\ref{eq:tau_af}, Eq.\ref{eq:prelu_alpha_case1}, Eq.\ref{eq:prelu_alpha_case2}, we divide Eq.~\ref{eq:rpelu_kappa} into four cases. (We exclude the trivial scenario of $\tau = 0$)

Case $1$: $\tau < 0$ and $b_1 \leq 0$

\begin{equation}
\begin{aligned}
\bA_b = \begin{cases}
1 & \bA_f < -b_1 / {\alpha \tau} - b_0 \\
-1 & \bA_f \ge -b_1 / {\alpha \tau} - b_0 \\
\end{cases}
\label{eq:tau_b1_case1}
\end{aligned}
\end{equation}

Case $2$: $\tau < 0$ and $b_1 > 0$

\begin{equation}
\begin{aligned}
\bA_b = \begin{cases}
1 & \bA_f < -b_1 / \tau - b_0 \\
-1 & \bA_f \ge -b_1 / \tau - b_0 \\
\end{cases}
\label{eq:tau_b1_case2}
\end{aligned}
\end{equation}

Case $3$: $\tau > 0$ and $b_1 \leq 0$

\begin{equation}
\begin{aligned}
\bA_b = \begin{cases}
1 & \bA_f > -b_1 / {\alpha \tau} - b_0 \\
-1 & \bA_f \leq -b_1 / {\alpha \tau} - b_0 \\
\end{cases}
\label{eq:tau_b1_case3}
\end{aligned}
\end{equation}

Case $4$: $\tau > 0$ and $b_1 > 0$

\begin{equation}
\begin{aligned}
\bA_b = \begin{cases}
1 & \bA_f > -b_1 / \tau - b_0 \\
-1 & \bA_f \leq -b_1 / \tau - b_0 \\
\end{cases}
\label{eq:tau_b1_case4}
\end{aligned}
\end{equation}

We merge the cases all above as:

\begin{equation}
\begin{aligned}
\bA_b = \begin{cases}
Sign(\tau)Sign(\bA_f + b_0 + b_1 / {\alpha \tau}) & b_1 \leq 0 \\
Sign(\tau)Sign(\bA_f + b_0 + b_1 / \tau) & b_1 > 0 \\
\end{cases}
\end{aligned}
\label{eq: ab_sign}
\end{equation}

So, the $\theta(n)$ in Eq.~\ref{eq:binary_A} can be solved with:

\begin{equation}
\begin{aligned}
\theta = \begin{cases}
-b_1 / {\alpha \tau} - b_0 & b_1 \leq 0 \\
-b_1 / \tau - b_0 & b_1 > 0 \\
\end{cases}
\end{aligned}
\label{eq: theta_solution}
\end{equation}

\section{Area and Energy} \label{sec:apdix_area_energy}

For simplicity, we define 8-bit systolic array as $16B \times 16B$, and the corresponding 1-bit systolic array as $128b \times 128b$. Let the area of systolic array be A.

For 8-bit systolic array, there are a multiplier and a adder in each Process Element(PE). Under the design of TSMC 7nm, the area of the multiplier is about 23 $\mu m^2$, and the area of the adder is about 14 $\mu m^2$. Then, the total area of int-8 array can be approximately computed as: (Because there are slacks in chip placement, so the area sum here can only be a lower bound in a 2D chip floorplan. Same below)
\begin{equation}
A_8(pe) = (23 + 14) \times 16 \times 16 = 9472 \mu m^2
\end{equation}
Additionally, a partial sum memory ($Psum_{mem}$) is needed for each column of PE to store and compute partial sum cyclically, and get the sum finally. There are 16 $Psum_{mem}$ for $16B \times 16B$ array. Following the common setting of 8-bit arrays, the bit width of partial sum is 32, and the depth used to store the middle results of partial sum is 1024. Under the design of TSMC 7nm, the area is about $2300 \mu m^2$ for each $Psum_{mem}$ of $1024 \times 32b$. So, the total memory of $Psum_{mem}$ is:
\begin{equation}
A_8(Psum_{mem}) = 2300 \times 16 = 36800 \mu m^2
\end{equation}

The total memory of the 8-bit systolic array is:
\begin{equation}
A_8 = 9472 + 36800 = 46272 \mu m^2
\end{equation}

There are an XNOR gate for each PE, and a POPCOUNT logit unit for each column for sum in a binary systolic array. Under the design of TSMC 7nm, the area of XNOR gate and POPCOUNT unit are 0.6  and 100 $\mu m^2$ respectively. So the total area of binary PE can be computed as:
\begin{equation}
A_{bnn}(pe) = 0.6 \times 128 \times 128 + 100 \times 128 = 22631 \mu m^2
\end{equation}

Also, the bit width of partial sum is 16 in binary systolic array. and the area is about $1400 \mu m^2$ for each $Psum_{mem}$ of $1024 \times 16b$. So, the total memory of $Psum_{mem}$ in binary systolic array is:
\begin{equation}
A_{bnn}(Psum_{mem}) = 1400 \times 128 = 179200 \mu m^2
\end{equation}

The area of array registers is small, and we ignore it. The total memory of the 8-bit systolic array is:
\begin{equation}
A_{bnn} = 22631 + 179200 = 201831 \mu m^2
\end{equation}

The ratios of $PEs$, $Psum_{mem}$ and array area are  2.4, 4.87 and 4.36 respectively. Decreasing the bit width of partial sum is a good way to reduce the area of binary systolic arrays.

The main power consumption of network inference occurs in convolution layers, and the contribution to power consumption of $Psum_{mem}$ is limited. The power consumption of convolution is proportional to the area of PEs. So, we can roughly estimate the power consumption of 1-bit systolic array is 2.5-4.36 times that of 8-bit, and much more close to 2.4 times.

\section{Convolution Cycle Formulae} \label{sec:cycle_formula}

Follow the parameter definition in Sec.~\ref{sec:cycle}, we will derive the cycle counts for a convolution operation. For the computation of a convolution layer, the total cycles consist of loading instructions, MAC computation, and writing the results. When the output bit-depth is less or equal to the input bit-depth, the time for writing results can be absorbed via pipelines. Higher output bit-depth will lead to data blocking. The number of instructions required by a convolution layer, denoted by $I$, is proportional to the the size of aligned output feature map. The cost cycle of each cycle is equal to array size $S$.
then,
\begin{equation}
I = \lceil \frac{N}{S} \rceil  \lceil \frac{W H}{P} \rceil
\end{equation}

which derives Eq.~\ref{instruction_cost}. When the input and output bit-depth are all 1-bit, the cycle count formula is described in Eq.~\ref{conv_cycle}. We define the number of cycles needed to write all the pixels in output feature map as Round($R$),

\begin{equation}
R = K \lceil \frac{CK}{S} \rceil
\end{equation}

When the output bit-depth is larger than 1 in binary array, writing of the last round of feature map is congested, and the number of cycles is proportional to the output bit-depth $b$. So the cycles in this case is described in Eq.~\ref{bnn_conv_1imo}.

\end{appendices}

\end{document}